\documentclass{article}
 
\usepackage[final]{neurips_2019}

\usepackage[utf8]{inputenc} 
\usepackage[T1]{fontenc}    
\usepackage{hyperref}       
\usepackage{url}            
\usepackage{booktabs}       
\usepackage{amsfonts}       
\usepackage{nicefrac}       
\usepackage{microtype}      
\usepackage{graphicx}
\usepackage{booktabs} 
\usepackage{epsfig,epstopdf}
\usepackage{algpseudocode}
\usepackage{balance}
\usepackage{amsmath,amssymb,amsfonts}
\usepackage{multicol,multirow}
\usepackage{mathrsfs}
\usepackage{eufrak}
\usepackage{color, graphics,graphicx}
\usepackage{algorithm}
\usepackage{threeparttable}
\usepackage{subfigure,caption}
\usepackage{cancel}
\usepackage{enumitem}

\def\ie{\emph{i.e., }}
\def\eg{\emph{e.g., }}

\def\yifan{\textcolor{black}}
\def\normaljudy{\textcolor{black}} 
\def\polishjudy{\textcolor{black}}

\title{Collaborative Unsupervised Domain Adaptation for \\Medical Image Diagnosis}

\author{
	Yifan Zhang$^{1,2}$, Ying Wei$^2$, Peilin Zhao$^{1}$, Shuaicheng Niu$^{1}$,  Qingyao Wu$^{1*}$,\\ \textbf{Mingkui Tan}$^{1}$\thanks{Corresponding author.\vspace{-0.15in}}  \textbf{,} \textbf{Junzhou Huang}$^{2}$  \\
	$^1$South China University of Technology, $^2$Tencent AI Lab \\
	sezyifan@mail.scut.edu.cn, $\{$qyw,mingkuitan$\}$@scut.edu.cn
}

\begin{document}

\maketitle

\begin{abstract} 

Deep learning based medical image diagnosis has shown great potential in clinical medicine. However, it often suffers two major difficulties in practice: 1) only limited labeled samples are available due to expensive annotation costs over medical images; 2)   labeled images may contain  considerable label noises (\eg mislabeling labels) due to  diagnostic difficulties. In this paper, we seek to exploit rich labeled data from relevant domains to help the learning in the target task with unsupervised domain adaptation (UDA).  Unlike most existing UDA methods which rely on clean labeled data or assume samples are equally transferable, we propose a novel Collaborative Unsupervised Domain Adaptation algorithm to conduct transferability-aware domain adaptation and conquer  label noise in a cooperative way. Promising empirical results verify the superiority of the proposed method.
\end{abstract}
 
%

\section{Introduction}
\label{sec:intro}

Deep learning has achieved great success in various real-world applications. One common prerequisite is rich annotated data~\cite{ref_partial1}. In medical image diagnosis, however, such \polishjudy{extensive} supervision is often absent due to \polishjudy{prohibitive} costs of data labeling~\cite{ref_survey2,ref_survey}, which impedes the successful application of deep learning. Hence, there is a strong motivation to develop unsupervised domain adaptation  (UDA)~\cite{ref_domain_miccai,ref_transfer} \polishjudy{methods to improve diagnostic accuracy with limited annotated medical images.}
 
In addition to the discrepancy between domains which all UDA methods~\cite{ref_partial1,ref_cada,ref_res,ref_mcd} aim to resolve, medical image diagnosis poses two additional challenges.
\textbf{First}, there also exist significant discrepancies among images within the same domain.
The discrepancy, such as in appearance and scale of the lesion region,
mainly arises from inconsistent data preparation procedures~\cite{ref_ours}, such as tissue collection, sectioning and staining.
In consequence, the difficulty of domain alignment  varies from sample to sample.
That is, the target samples which bear a striking similarity with source samples are easier to align than the (hard-to-transfer) samples that are highly dissimilar. 
Existing UDA methods~\cite{ref_DSN,ref_DANN,ref_ADDA,ref_DDC}, however, often ignore such sample differences and treat all data equally. As a result, some hard-to-transfer images may not be well treated, leading to inferior overall performance. 
\textbf{Second}, a considerable percentage of medical annotations are unfortunately noisy, \yifan{caused by} diagnostic difficulties~\cite{ref_survey2}. Essentially, noisy labels are corrupted from ground-truth labels and thus degrades performance of learned classifiers on the clean data distribution~\cite{ref_cotraining}.
As a result, directly applying the classifier built upon noisy source data inevitably performs poorly on the target domain, even though it has been aligned well with the source.
 
\begin{figure}[t]
\centering
\vspace{-0.3in}
\includegraphics[width=12.5cm]{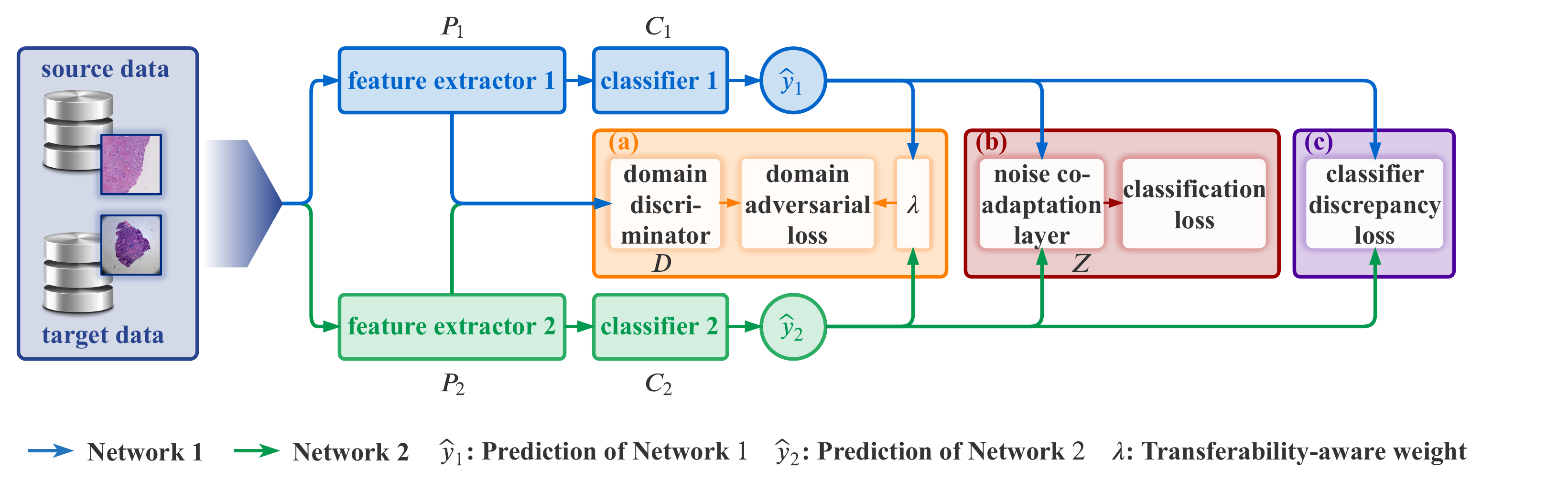}
\vspace{-0.05in}
\caption{The scheme of CoUDA, where $P$ denotes feature extractors, $C$ means classifiers, and $Z$ is noise co-adaptation layer. The final prediction is the average of two networks' predictions $\bar{y} \small{=} \frac{\hat{y}_1+\hat{y}_2}{2}$.} \label{framework}
\vspace{-0.2in}
\end{figure}

To address the two challenges that are usually ignored by existing UDA methods~\cite{ref_DANN,ref_ADDA}, we  propose a Collaborative Unsupervised Domain Adaptation (CoUDA) algorithm. As shown in \textbf{Figure~\ref{framework}}, by taking advantage of the collective intelligence of two (or possibly more) peer networks,
CoUDA is able to distinguish between samples with different levels of domain alignment difficulty  (\textbf{module (a)}).
Then, CoUDA assigns higher importance to the hard-to-transfer examples, which are detected by peer networks  with greater prediction inconsistency.
Meanwhile, we overcome label noise by a novel noise co-adaptation layer, shared between peer networks  (\textbf{module (b)}).
The layer aggregates different sets of noisy samples identified by all peer networks, and then denoises by adapting the predictions of these noisy samples to their noisy annotations. To further maximize the collective intelligence, we enforce the classifiers of peer networks to be diverse as large as possible  (\textbf{module (c)}). In this way, we expect that the cooperative network diagnoses target medical images  better.

\vspace{-0.1in}
\section{Methods}  
\vspace{-0.05in}
\textbf{Problem definition:} we focus on the issue of unsupervised domain adaptation, where the model has access to only unlabeled target images.
We denote $\mathcal{S}\small{=}\{x_i^s, y_i^s\}_{i=1}^{n_s}$ as the source domain of $n_s$ samples, where $x_i^s$ denotes the $i$-th sample and  $y_i^s \small{\in} \{0,1\}^K$ denotes its label with $K$ being the number of classes.
In practice, we may only observe source images with noisy labels, and we denote them as $\mathcal{S}'\small{=}\{x_i^s, z_i^s\}_{i=1}^{n_s}$.
The unlabeled  target domain
$\mathcal{T}\small{=}\{x_j^t\}_{j=1}^{n_t}$ can be defined in a similar way.

\textbf{Collaborative domain adaptation}: 
Intuitively, hard-to-transfer samples are often difficult to be correctly classified by two networks simultaneously~\cite{ref_mcd}. 
Since two peer networks generate decision boundaries with different discriminability, we propose to detect data transferability based on prediction inconsistency of two networks. 
Specifically, we define the transferability-aware weights as: $\lambda \small{=} 1 \small{-}\cos{(\hat{y}_{1},\hat{y}_{2})}$, where $\hat{y}_{1}$ and $\hat{y}_{2}$ are the prediction probabilities by two classifiers and the cosine distance  measures the prediction similarity. 
Based on this weight, we conduct transferability-aware domain adaptation in an adversarial manner~\cite{ref_GAN,cao2019multi}.
Specifically, we define the domain adversarial loss as:
$\mathcal{L}^{adv} \small{=} \sum_{\tau} \Big{[}\frac{1}{n_t} \sum_{\normaljudy{\mathcal{T}}}\lambda_j^t \big{(}p(\hat{d}_\tau\big{|}x_j^t)-1\big{)^2} + \frac{1}{n_s}\sum_{\mathcal{S}}\lambda_i^s \big{(}p(\hat{d}_\tau\big{|}x_i^s)\big{)^2}\Big{]},  $
where $\tau
\small{\in}\{1,2\}$ indicates the index of peer networks. Here, $\lambda_j^t$ and $\lambda_i^s$ denote the transferability-aware weights, and   $\hat{d}$ denotes the predicted   domain probability by the shared domain discriminator $D$.

\textbf{Collaborative Noise adaptation:}  
To make deep neural networks more robust, we follow the class-conditioned noise assumption~\cite{ref_noiselayer} that the noisy label $z$ is conditioned on both the true label $y$ and data features $f$, \ie $p(z|y,f)$.
As a result, the prediction regarding to the mislabeled class $m$ can be expressed as $p(z\small{=}m|x,f)\small{=}\sum_{k=1}^K p(z\small{=}m|y\small{=}k,f) p(y\small{=}k|x)$, where $p(z\small{=}m|y\small{=}k,f)$ means the probability that the true label $k$ changes to noisy label $m$. Based on this assumption, we propose a noise co-adaptation layer $Z$ to estimates the noise transition probability using the softmax function:
$
 p(\hat{z}\small{=}m|\hat{y}\small{=}k,f)\small{=}\frac{\exp{\big{(}{w^{\top}_{km}f+b_{km}\big{)}}}}{\sum_{l=1}^K \exp{\big{(}{w^{\top}_{kl}f+b_{kl}\big{)}}}},
$
where $\hat{z}$ denotes the prediction of the noisy label and $f$ represents the learned features by feature extractors. Moreover, $w_{km}$ and $b_{km}$ indicate the parameters of $Z$ regarding to true label $k$ changing to noisy label $m$.
Here, the noise co-adaptation layer is shared by two peer networks, since different networks have different discriminative abilities and thus have different abilities to filter label noise out~\cite{ref_mcd}. That is, they are able to adjust the estimation error of noise transition probabilities, possibly caused by a self-evolving network.
Following this, we predict the noisy label by:
$p(\hat{z}\small{=}m|x,f)\small{=} \sum_k p(\hat{z}\small{=}m|\hat{y}\small{=}k,f)  p(\hat{y}\small{=}k|x)$, and train peer networks by matching the adapted predictions to noisy labels. Since the class imbalance~\cite{zhang2018online,zhang2020online,zhao2018adaptive} is very common in medical image diagnosis, we adopt the focal loss~\cite{ref_Focal} as our classification loss $\mathcal{L}^c$. 

\textbf{Classifier Diversity Maximization:} %
Note that the detection of data transferability highly depends on the classifier diversity. Also, keeping classifiers diverse prevents the noise co-adaptation layer reducing to a single noise adaptation layer~\cite{ref_noiselayer} in function. Hence, we further  maximize the classifier discrepancy  based on Jensen-Shannon (JS) divergence~\cite{ref_GAN}:
$ \mathcal{L}^{dis} \small{=} D_{JS}(\hat{y}_{1}\|\hat{y}_{2})$.

\textbf{Overall Training Procedure} 
 is to solve the following minimax problem: 
$   \min_{P_{\tau},C_{\tau},Z}  \max_{D} \mathcal{L}^c - \alpha \mathcal{L}^{adv}    \small{-} \eta\mathcal{L}^{dis}$,
where $\tau
\small{\in}\{1,2\}$ is the index of peer networks, while $\alpha,\eta$ are hyper-parameters.  


\section{Experimental Results}\label{experiment}
 
\textbf{Dataset}: The medical image dataset is formed by H$\&$E stained colon histopathology slides, which are diagnosed as three types of colon polyps (normal, adenoma, and adenocarcinoma). The details of data acquisition can be found in~\cite{ref_ours}, while the data statistics are shown in Table~\ref{dataset}. Here, the \textbf{whole slide image} (WSI) is regarded as the labeled source data, while the \textbf{microscopy image} (MSI) is viewed as the unlabeled target data. 
As shown in Table~\ref{dataset}, the class imbalance  is quite severe.
Moreover, according to the remark of doctor annotators, there supposedly exist a certain number of noisy labels. Considering both class imbalance and label noises, this problem is very tough.

\begin{table}[h]
    \vspace{-0.15in}
	\caption{Statistics of medical image dataset.}\label{dataset}
     \begin{footnotesize}
    \begin{center}  
    \scalebox{0.8}{
	\begin{tabular}{|c|c|c|c|c|c|}\hline
        \multirow{2}{*}{Set} &\multirow{2}{*}{Domain} &\multicolumn{3}{c|}{Categories}&\multirow{2}{*}{total} \\    \cline{3-5} 
          & & normal& adenoma& adenocarcinoma  & \cr 
        
        \hline 
        \multirow{2}{*}{Training} &
        WSI (source domain)         &36,094 	&3,626 		& 3,081 	 & 44,542  \\
        &MSI (target domain)  &2,696  	& 1,042	 	& 1,091 		& 7,363	\\ 
        \hline
        Test &
        MSI (target domain)  	& 1,110 	& 487 & 713   & 2,984	\\
        \hline
	\end{tabular}}
    \end{center}
    \end{footnotesize} \vspace{-0.15in}
\end{table}

\textbf{Results}:
We report the results in Table~\ref{performance}.
\textbf{Overall}, our proposed method achieves the best performance, which confirms its superiority in medical image diagnosis. Since there is an urgent need for robust deep models for medical image diagnosis, CoUDA makes great clinical sense in practice.  In comparison: 
\textbf{First}, MobileNet~\cite{ref_Mobilenet} (the backbone network for all methods)  cannot learn a model with high generalization ability, since it is unable to solve the issues of domain discrepancies and label noise.
\textbf{Second}, Co-teaching~\cite{ref_coteaching} and NAL~\cite{ref_noiselayer} (classic weakly-supervised learning methods) perform better than MobileNet, since they are able to address label noise to some degree. However, they do not consider the discrepancy between domains and hence cannot adapt domains.
\textbf{Third}, DDC, DANN and MCD (classic unsupervised domain adaptation methods) also perform well, which confirms the necessity of domain adaptation. However, they assume samples have clean labels and also ignore the data difference in transferability, thus leading to insufficient performance.  

\begin{table}[h] 
    \vspace{-0.15in}
	\caption{Comparisons on medical image diagnosis in terms of four metrics.}
	\label{performance} 
\begin{footnotesize}
 \begin{center} 
 \scalebox{0.8}{
 	\begin{tabular}{|l|c|c|c|c|}\hline
        Methods& Accuracy (\%) & Macro Precision & Macro Recall  & Macro F1-measure   \cr
        \hline 
         MobileNet~\cite{ref_Mobilenet}  &68.79	 	& 78.62	& 61.67	& 64.46        \\
         Co-teaching~\cite{ref_coteaching}  & 71.52 &79.74	& 64.88	& 67.93       \\	
         NAL~\cite{ref_noiselayer}  &68.92 	& 74.83	& 63.49	& 64.65               \\ 
         \hline 
         DDC~\cite{ref_DDC} & 79.74 & 78.75	& 79.01 		& 78.76	           \\
         DANN~\cite{ref_DANN} & 80.01 	& 79.25	& 78.85 	& 78.75              \\
         MCD~\cite{ref_mcd} &81.77 	& 81.04	& 80.45	& 80.19                   \\
          CoUDA (ours)  & \textbf{87.75} 	& \textbf{87.62}	& \textbf{86.85} 	& \textbf{87.22}      \\
        \hline
 
	\end{tabular}}
	 \end{center} 
	  \end{footnotesize} 
	  
	  \vspace{-0.1in}
\end{table}

\textbf{Ablation Studies}:
We then conduct ablation studies. As shown in Table~\ref{Ablation}, all components in our methods (\ie collaborative scheme, three losses and noise co-adaptation layer), make empirical sense and play important roles in our method. In particular, domain adversarial loss with transferability-aware weights ($\mathcal{L}^{adv}$)  and noise co-adaptation layer (NCL) are relatively more important. This result verifies the necessity to reduce  domain discrepancies and overcome label noise.


\begin{table}[h] 
 \vspace{-0.1in}
	\caption{Ablation studies, where ``ours”   indicates the cooperative network,  NCL means the noise co-adaptation layer, and w/o denotes ``without”.}
	\label{Ablation} 
\begin{footnotesize}
 \begin{center} \scalebox{0.8}{
 	\begin{tabular}{|l|c|c|c|c|}\hline
        Methods& Accuracy (\%) & Macro Precision &  Macro Recall  & Macro F1-measure  \cr
        \hline  

         single peer network \small{+} $\mathcal{L}^c$ &68.92 	& 74.83	&63.49	& 64.65                    \\
         ours \small{+} $\mathcal{L}^c$  &72.51 	& 77.52	&66.65	& 69.24                \\
         ours  \small{+} $\mathcal{L}^c$ \small{+} $\mathcal{L}^{adv}$ &85.54 	& 84.49	&85.22	& 84.81 \\ 
         ours \small{+} $\mathcal{L}^c$ \small{+} $\mathcal{L}^{adv}$ \small{+} $\mathcal{L}^{dis}$ \small{+} w/o NCL &82.94 	& 81.37	&82.70	& 81.64    \\
          ours \small{+} $\mathcal{L}^c$ \small{+} $\mathcal{L}^{adv}$  \small{+} $\mathcal{L}^{dis}$  & \textbf{87.75} 	& \textbf{87.62}	& \textbf{86.85} 	& \textbf{87.22}        \\
  
        \hline
	\end{tabular}} 
	 \end{center} 
	  \end{footnotesize} 
	  \vspace{-0.15in}
\end{table}


\section{Conclusion} 
We have proposed a novel  Collaborative Unsupervised Domain Adaptation method for medical image diagnosis. Unlike previous UDA methods that treat all data equally or assume data with clean labels, our method cooperatively eliminates domain discrepancies with more focuses on hard-to-transfer samples, and overcome label noise simultaneously. 
Promising results on classifying real-world medical images demonstrate the superiority and effectiveness of the proposed method.
Since there is an urgent need for robust deep models for medical image diagnosis, our method is of great clinical significance in practice.
 
\newpage 
 
\section{Acknowledgement}
This work was partially supported by National Natural Science Foundation of China (NSFC) (61876208, 61502177 and 61602185), key project of NSFC (No. 61836003), Program for Guangdong Introducing Innovative and Enterpreneurial Teams 2017ZT07X183, Guangdong Provincial Scientific and Technological Funds (2017B090901008, 2017A010101011, 2017B090910005, 2018B010107001), Pearl River S$\&$T Nova Program of Guangzhou 201806010081, Tencent AI Lab Rhino-Bird Focused Research Program (No. JR201902), CCF-Tencent Open Research Fund RAGR20170105, Program for Guangdong Introducing Innovative and Enterpreneurial Teams 2017ZT07X183.

%
 
\medskip
 
\small
\bibliographystyle{abbrv}

\end{document}